\documentclass[10pt,twocolumn,letterpaper]{article}

\usepackage{cvpr}
\usepackage{times}
\usepackage{epsfig}
\usepackage{graphicx}
\usepackage{amsmath}
\usepackage{amssymb}

\usepackage{afterpage}
\usepackage{graphicx,float,subcaption}
\usepackage[ruled,vlined,linesnumbered]{algorithm2e}
\usepackage[toc,page]{appendix} % for appendices
\usepackage[breaklinks=true,bookmarks=false]{hyperref}
\usepackage[nameinlink, capitalise]{cleveref}
\usepackage[T1]{fontenc}
\usepackage{titling}
\usepackage[textwidth=18cm,textheight=23cm]{geometry}

\cvprfinalcopy

\def\cvprPaperID{****} % *** Enter the CVPR Paper ID here

\begin{document}

%%%%%%%%% TITLE
\title{\vspace{-1cm} \LARGE The Role Of Biology In Deep Learning}

\author{
Robert Bain\\
Oregon State University\\
{\tt\small bainro@oregonstate.edu}
}

\maketitle

\begin{abstract}
Artificial neural networks took a lot of inspiration from their biological counterparts in becoming our best machine perceptual systems. This work summarizes some of that history and incorporates modern theoretical neuroscience into experiments with artificial neural networks from the field of deep learning. Specifically, iterative magnitude pruning is used to train sparsely connected networks with 33x fewer weights without loss in performance. These are used to test and ultimately reject the hypothesis that weight sparsity alone improves image noise robustness. Recent work mitigated catastrophic forgetting using weight sparsity, activation sparsity, and active dendrite modeling. This paper replicates those findings, and extends the method to train convolutional neural networks on a more challenging continual learning task. The code is publicly available \href{https://github.com/bainro/active_dendrites}{\textcolor{blue}{here}}.
\end{abstract}

\section{Introduction}
\label{Introduction}

Sparsity has proven useful numerous times in deep learning (DL) \cite{l0_norm,dropout,sg_moe,switch_trans,lth,dense,beyond}. Unfortunately though, typical hardware does not effectively benefit from sparse weights (e.g. GPUs). FPGAs and neuromorphic hardware do however better leverage sparsity \cite{numenta_fpga, neuromorphic1, ratebased5, ratebased6}. The hardware lottery hypothesis is worth considering \cite{hardware_lott}. It is the idea that worse algorithms in AI can become popular because the current hardware favors them more, and not because they are better long term. This is supported by the fact that brains are sparse in both their connections and outputs.

\begin{figure}%[H]
  \centering
      \hspace*{0.85cm}
      \includegraphics[width=0.7\columnwidth]{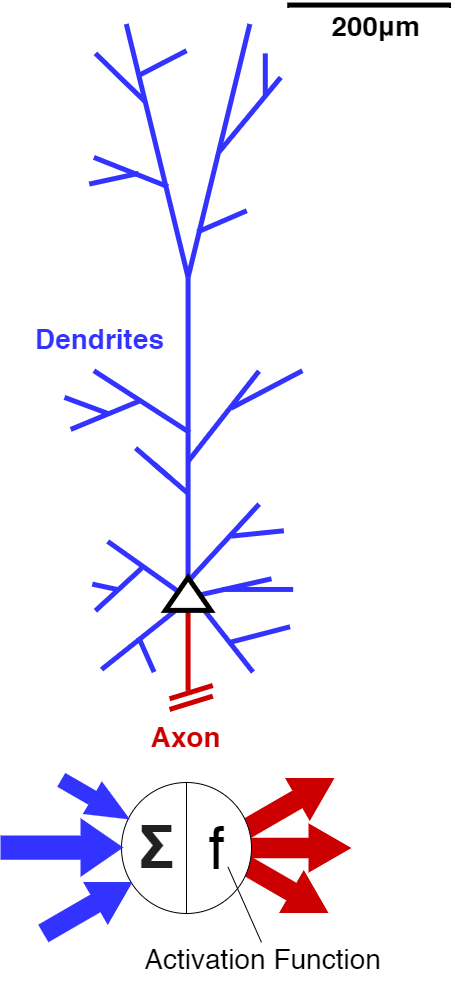}
      \caption{\textbf{Top}: Diagram of a human L5 pyramidal neuron. Dendrites receive information from other neurons in the form of voltage spikes. They transform the spikes while relaying them to where the cell body meets the axon. If the voltage across that bit of membrane sums to larger than a certain threshold value, a spike (i.e. action potential (AP)) is generated, and becomes input to other neurons. The axon is truncated as they are usually long and elaborate. \textbf{Bottom}: A point neuron, typically used in deep learning, abstracts away all physical structure of a real neuron. The influence a dendritic input has on an action potential being generated is modeled as a multiplicative weight. The dynamics at the soma are represented by a summation of weighted inputs, and an activation function. The latter is most often ReLU, which outputs 0 if the summed input is below 0. This in part represents the threshold of AP firing in real neurons.}
      \label{fig:real_deal}
\end{figure}

There is a lot of evidence that dendrites (see \cref{fig:real_deal}) are nonlinear filters of voltage signals as they travel to the cell body (i.e. soma) \cite{antic10, spruston08, chavlis_21}, but these active dendrite properties are not typically modeled in deep neural networks (DNNs). \cite{beyond} recently modeled some of these effects in what they called an \textit{Active Dendrites Network} (ADN). ADNs do well at PermutedMNIST, a continual learning task, by mitigating the effects of catastrophic forgetting. These were feedforward networks with few layers trained on a dataset that allowed for no transfer of knowledge between tasks. This paper  extends that work to include convolutional neural networks (CNNs) and tests it on a more challenging continual learning dataset that theoretically enables some knowledge transfer \cite{knowledge_trans}.

It is worth reflecting on the history of artificial intelligence (AI) research to study what has worked the best in the past. This work attempts to follow a rather long tradition of using data from biological sciences to create better AI. Part of that history is chronicled in the subsequent \textit{\cref{dl_back}}.

\setlength{\belowcaptionskip}{-10pt}
\setlength{\belowcaptionskip}{0pt}

\section{Background}
\label{Background}
%%%
\subsection{Deep Learning Background}
\label{dl_back}
%%%

 McCulloch and Pitts \cite{M_and_Pitts} created the point neuron model most commonly used in deep learning today\footnote{Lapique is often incorrectly cited, but did nearly invent the modern leaky integrate and fire (LI\&F) neuron model \cite{lapiqueTrans, lapique2008}. The seemingly only missing component would be a variable resistor to model the cell membrane gates as they open and close.}. They also proposed that biological neural networks (BNNs) were equivalent to binary first order logic and could be implemented in electronic circuits. \cite{Hebb} improved on this by allowing unequal (ie weighted) connection strengths between neurons. Rosenblatt's perceptron \cite{Perceptron} tested Hebb's idea that modifying such connection strengths could lead to memory and learning, to great success. Minsky \& Papert (1969) are cited as causing the subsequent AI winter by writing that perceptrons could not even learn to do a trivial XOR on 2 inputs. Solutions were already known at the time however \cite{RosenblattBook}, by including multiple layers of neurons and non-linear activation functions, but the damage to the perceptron's reputation had already been done by the point Minsky and Papert became aware of such solutions. 

Fukushima created convolutional layers with multiple input and output feature maps, downsampling, and weight sharing (ensuring translational invariance) with the neocognitron in 1980 \cite{neocog}. 

Lecun's architecture years later was largely the same, but with back-propagation of errors (i.e. backprop) to change weights instead of self-supervised learning \cite{lecun_tied, lecun85}. Lecun cites Rumelhart's T-C problem paper for weight sharing \cite{backprop}, which in turn cites Fukushima's neocognitron for using an analogous solution. 

All of the aforementioned scenarios involved low data and compute regimes. Lecun was dealing with a dataset of only 480 tiny 16x16 pixel grayscale images. Weight sharing was demonstrated to be compute efficient since less parameters had to be learned, and useful to avoid overfitting small datasets. 

Fukushima took direct inspiration from Hubel and Weisel's work on the striate cortex (i.e. V1) of cats\footnote{They euthanized 40 cats. Bring that up in your AI ethics course.} and monkeys \cite{cat_v1, monkey_v1}. The neocognitron's S \& C cells correspond to simple \& complex cells from that work. The nomenclature of \textit{receptive fields} is also borrowed from the same literature. That term is still used heavily today, but many artificial neural network (ANN) practitioners do not realize it comes from investigations into the eyes and brains of animals.

ReLU came about somewhere between Rosenblatt's 1962 book "\textit{Principles of Neurodynamics},"\cite{RosenblattBook} and Fukushima's 1968 paper about "analog threshold units" \cite{Fukushima68}. The sub-sections of Ch. 10 \cite{RosenblattBook} involve transfer functions that each have components of ReLU as used today (e.g. thresholding, non-linearity, and monotonically increasing functions). The neocognitron paper's figure 3 includes the exact input-output response plot we typically see today when learning about ReLU.

The deep CNNs used today bear a great deal of similarity to Fukushima's Neocognitron. Perhaps the biggest software difference is the greater depth, which is attainable through residual (i.e. skip) connections \cite{resnet}. Alexnet \cite{alexnet} is often credited as having started the latest AI summer in 2012 by introducing ReLU activation functions and leveraging the massively parallel architecture of GPUs. ReLU is quick to calculate in both the forward and backward passes, and blocks certain neurons from contributing to the next layers' neuron's firing. This reduces training time, and induces sparser weight updates respectively. As noted earlier, ReLU was already applied in ANNs during the 1960s, which leaves just hardware changes to credit for the latest wave of interest in ANNs. While the brain is massively parallel like GPUs, the brain is sparse in its connections, connection updates, and activations (i.e. the percentage of neurons sending action potentials at any given moment). GPUs currently do not enable speedups when doing large sparse matrix calculations like those required for DNNs. The performance benefits are usually only double, even in cases of rather extreme weight and activation sparsity.

%%%
\subsection{Active Dendrites Background}
%%%

% \begin{figure}
%   \centering
%       \hspace*{-1.4cm}
%       \includegraphics[width=0.88\columnwidth]{images/boat_pyramidal.png}
%       \caption{\textbf{Left}: Diagram of L5 pyramidal neuron highlighting the distinct dendritic zones. Each can have different connection learning rules, perform different calculations, and interact in complex ways. \textbf{Bottom}: Dashed green overlays show where NMDA spikes can occur in typical L5 pyramidal neurons. Connections roughly within the blue circle are close enough to the cell body to have a large influence on whether an AP is produced, and does not require active dendritic effects to do so.}
%       \label{fig:boat}
% \end{figure}

Dendrites have been modeled as passive voltage attenuators for a long time. The majority of excitatory inputs (~85\%) on L5 pyramidal cells are on dendrites frequently too far away from the cell body to significantly contribute to axonal APs without active dendritic properties \cite{larkum91, major13, spruston08, chavlis_21, jarsky06}. These properties include at least sodium (Na+), potassium (K+), calcium (Ca++), NMDA, and back-propagating action potential activated Ca++ (BAC) spikes.

A lot of discoveries have been been made since the 1990s about NMDA spikes in the dendrites, largely in part due to better realtime imaging \cite{antic10}. The NMDA channel membrane protein that allows such spikes to occur requires both glutamate molecules to bind the receptor, and high enough voltage levels to evacuate magnesium from the channel's pore. Even with high transmembrane voltages, NMDA spikes do not spread beyond where glutamate is bound. This local type of spike is in contrast to the typical Na+ spikes that travel along axons just through voltage gradients alone. NMDA spikes occur primarily in the thin, distant dendrites, but not in the apical trunk (the black bit of the dendrites in \cref{context_gating}). Dendritic branches typically have 100-400 spines. NMDA spikes are highly localized events that envelop just one small dendritic segment, 10 – 40 $\mu$m in length \cite{antic10}. As few as 8 individual inputs from other neurons can produce NMDA spikes if they are close enough to each other on the dendrite. NMDA spikes over larger dendritic surface areas require stronger or more concurrent inputs \cite{spruston08, antic10}.

Point neurons assume that the axonal trigger zone is the only place where spiking occurs. This assumption was made a long time ago, and is no longer assumed to be true by electrophysiologists. Voltage sensitive Na+, K+, and Ca++ channels are embedded in the membrane nearly everywhere, and assist in actively propagating inputs and dendritic spikes towards the soma, as well as axonal APs back into the dendritic arbor \cite{magee_95, waters2015, stuart_94, spruston_95}. The latter signals are called backwards propagating action potentials, or bAPs (no relation to backprop in DL). The filtering of bAPs can be nonlinearly conditioned on previous neuron activity, as voltage levels can cause bAPs to be attenuated differently \cite{ledergerber10, spruston08}.

NMDA spikes in the dendrites most proximal to the cell body of L5 pyramidal cells display rather binary voltage responses wrt input intensity, but the relationship between intensity and duration is positively linear \cite{antic10}. Stronger inputs can thus lead to more prolonged NMDA spikes, which is thought to be useful for integrating information from multiple modalities (i.e. sensor fusion).

It is known today that the apical trunk of pyramidal neurons alone can do XOR \cite{dCaAPs} via active dendritic processing. Even relatively small compartments of a single biological neuron can compute XOR, where point neurons typically require 2 layers of processing to perform XOR.

Small dendritic spikes often need to sum with other inputs to evoke spiking at the soma, or to make it more likely to fire. Dendritic spike interaction also enables complicated coincidence detection \cite{spruston08, antic10}. 

Many researchers have highlighted the importance of active dendrites in updating connection strengths between biological neurons \cite{waters2015, antic10, spruston08, chavlis_21}. Blocking bAPs through chemical means can prevent local updates rules from occurring within pyramidal cells \cite{stdp1, stdp2}.

%%%
\subsection{Lottery Ticket Hypothesis Background}
%%%

The lottery ticket hypothesis (LTH) \cite{lth} states that densely connected DNNs contain sparse sub-networks that can exceed the test accuracy of the original network after training on at most the same number of instances. These sparse networks are called winning lottery tickets (WLTs) and do much better than the average random sub-network.

\begin{algorithm}
  \SetAlgoLined
\textcolor{white}{;} Train model to convergence\\\textcolor{white}{;}
Set x\% of the smallest magnitude, non-zero weights per layer to 0.\\\textcolor{white}{;}
Set non-zero weights to their original values before training.\\\textcolor{white}{;}
Train model to convergence again\\\textcolor{white}{;}
Repeat steps 2 through 4  until the desired sparsity is met.
  \caption{Iterative Magnitude Pruning}
  \label{pseudocode_imp}
\end{algorithm}

\cite{lth} used iterative magnitude pruning (see \cref{pseudocode_imp}) to create WLTs. They have been discovered in many architectures \cite{uber, lth, otra_domains_rl_nlp, rejected}, and in different task domains like reinforcement learning (RL) and natural language processing \cite{otra_domains_rl_nlp}. Finding WLTs with more than 50\% sparsity appears trivial, but what pragmatic benefit do they offer to practitioners? Finding these winning tickets still requires a lot of compute and fine-tuning for larger CNNs. 

\cite{uber} found that even better than IMP is keeping weights whose values change by the greatest magnitude, instead of just keeping the largest magnitude weights. \cite{lth} unintentionally foreshadowed this when noting in their Appendix F that winning tickets' weights move further than other weights. \cite{uber} also showed that re-initializing the weights is not as important as retaining the original signs of the weights, lending even more evidence to the idea that re-initializing to the original values is not vital to finding WLTs.

\cite{dont_reinit} revised some of the experiments from \cite{lth} on pruning larger CNNs while training on ImageNet. This domain usually requires a more exhaustive hyperparameter search to discover WLTs, but when discovered the results tend to more impressive \cite{lth}. \cite{dont_reinit} demonstrate that using the more de facto training regime with a larger learning rate and momentum SGD instead of Adam produces better results than even the winning tickets, making LTH even less practical.

As originally noted by \cite{lth} their IMP method produces sparse networks in such a way that GPUs do not benefit. The sparsity allows more compression, but not faster inference. In part, this is due to the asynchronous checkpoints that occur while GPUs process matrix calculations in parallel. Multiplications by 0 in a certain batch might be faster (i.e. fewer clock cycles) but the slowest multiplication in that batch of operations has to finish before more calculations can be queued.

\section{Experiments}
\label{Experiments}

This \href{https://github.com/rahulvigneswaran/Lottery-Ticket-Hypothesis-in-Pytorch}{\textcolor{blue}{code}} was adapted to produce WLTs on MNIST and CIFAR10 (see \cref{pseudocode_imp}). Both datasets were trained on with fully connected (FC) networks, and the former also had a LeNet-5 trained on it. The best network from a max of 100 rounds of training was kept, then 12.5\% of the remaining weights were pruned. This train-prune loop occurred until only 1.1\% of the original weights were left being non-zero. The same networks were randomly pruned as well. The resulting WLTs and random tickets were then tested for noise robustness by adding variable amounts of noise to the test set images. Examples from that process can be visualized in \cref{fig:lth_train}. All trends reported are an average over 8 training runs with different seeds.

\begin{figure}
        \includegraphics[width=\columnwidth]{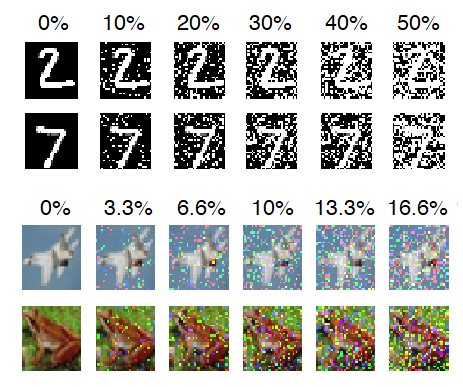}
        \caption{Examples of the images used to test the effects of weight sparsity on noise robustness. The probability of injecting noise was uniform across the image, but the noise was created by setting a pixel's value to near full intensity (specifically 2 standard deviations above the dataset's mean). \textbf{Top}: Examples from MNIST. \textbf{Bottom}: Examples from CIFAR10 required less noise for me to have difficulty classifying them.}
        \label{fig:noise_viz}
\end{figure}

\begin{figure*}
  \centering
  \includegraphics[width=0.8\textwidth]{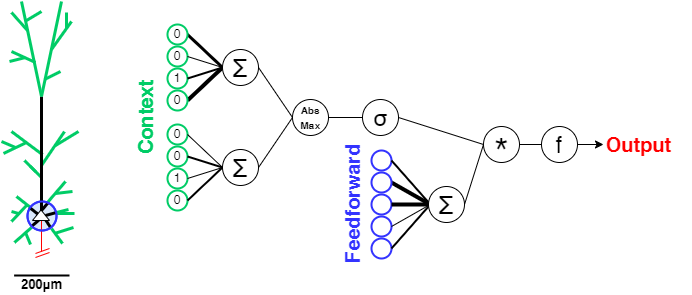}
  \caption{\textbf{Left}: Diagram of L5 pyramidal neuron highlighting dendrites that produce NMDA spikes in green. The blue circle encompasses the connections close enough to the cell body to have a large influence on producing APs without active dendritic effects. \textbf{Right}: A diagram of a single active dendrite neuron model. This particular one has 2 dendritic compartments. Each gets a copy of the context signal. As shown this would be a network trained on 4 consecutive tasks, the current being the 3rd, as indicated by the 1-hot encoding scheme. \cite{beyond} used the task averaged input as context, leading to networks with many more weights. The absolute max is taken among dendritic segments, and then scaled between 0 and 1 using a sigmoid function. \textit{f} stands for the chosen activation function. If it is kWTA, then the layer's other neuron's values at this point in the diagram will be taken into consideration as well.}
  \label{context_gating}
\end{figure*}

\begin{figure}
  \includegraphics[width=\columnwidth]{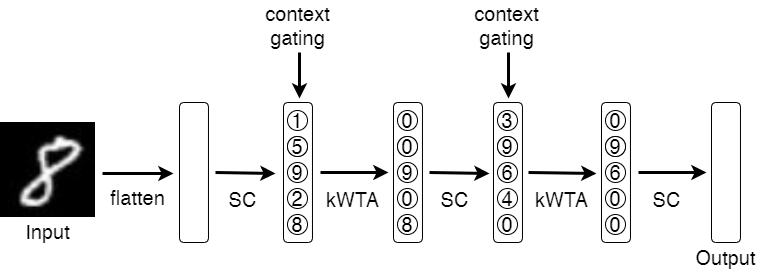}
  \caption{An ADN network architecture similar to those used in \cite{beyond}. SC stands for sparsely connected feedforward layer, and kWTA stands for a \textit{k-winners take all} activation function. Note that k=2 for the shown kWTA activation functions (i.e. 40\% sparse).}
  \label{adn_pmnist}
\end{figure}

Next, I set out to replicate some of the continual learning results on PermutedMNIST reported by \cite{beyond}. PermutedMNIST \cite{pmnist} is a continual learning benchmark that takes normal MNIST and applies a random shuffling of the pixels (i.e. permutes the pixels), but does the same shuffling across each task. Thus many tasks can be made from just MNIST, but there is no semantic information that can be leveraged across individual tasks.

\begin{equation}
%\begin{align*}
    y =\left(\boldsymbol{w}^{\top} \boldsymbol{x}+b\right) \times \sigma\left(\max _{j} \boldsymbol{u}_{j}^{\top} \boldsymbol{c}\right)
%\end{align*}
\label{eq:single_n}
\end{equation}

\cref{eq:single_n} details how the active dendrite model calculates a neuron's output ($y$) using context ($\boldsymbol{c}$) and feedforward inputs ($\boldsymbol{x}$). There are j sets of dendritic segments, each with a set of weights ($\boldsymbol{u}$). Every segment takes a weighted sum of the context vector. The largest absolute value among segments is passed through a sigmoid, and is then multiplied by the feedforward output of the neuron. This is referred to as context gating. See \cref{context_gating} for a visualization of this equation.
 
Their ADNs are detailed in \cref{adn_pmnist}, which are built upon layers of active dendrite neurons as shown in \cref{context_gating}. They used the element-wise average image as the context signal to each ADN neuron. In \cite{numentas_latest_RL} they used 1-hot encoding instead, which we tried out as well to see the effects of fewer context weights. Every task gets its own unique 1-hot encoded vector in this setting. To further explore the effect of more context weights, we also trained the 1-hot ADN architectures with a context vector full of 0s. In this setting there is effectively no task-specific information given as context, just a single multiplicative gating bias for certain layers' neurons, each of which is calculated from multi-layer perceptrons (MLPs) given 0s as input. For the 1-hot and 0s context vectors, we used the hyperparameters \cite{beyond} found from their grid searches.

To induce activation sparsity they used k-winners take all (kWTA), which can be seen as a ReLU with a variable threshold \cite{kwa}. The threshold is set at each inference such that k neurons are always active.

\begin{align*}
    k \mathrm{WTA}\left(y_{i}\right)= \begin{cases} y_{i}, & \text { if } y_{i} \text { is one of the largest k values in } y \\ 0, & \text { otherwise }\end{cases}
\end{align*}

\cite{beyond} did not track the test accuracy of the latest trained task while doing continual learning, only the overall average of all tasks trained up to that point. I thought this worth doing to see if perhaps the larger network capacity has an advantage the whole time, as might be the case if it does better on all individual tasks.

All ADNs tested modeled 10 dendritic segments per active dendrite neuron. The number of inputs to each of those segments is determined by the size of the context vector, which is the same length for the 1-hot and 0s vector case (i.e. the number of tasks being trained on). All ADN results came from an average over 5 different seeds.

SplitCIFAR100, a new dataset, was made to test how sparse CNNs and ADNs generalize to more difficult tasks than they were tested in \cite{beyond}. It is a variant of SplitCIFAR from \cite{syn_int}, but only uses CIFAR100 and not CIFAR10. CIFAR100 has 100 classes typically, but here those classes are randomly sampled without replacement to create 10 sequential 10-way classification tasks. To extend ADNs to include CNNs the active dendrite modeling as multiplicative context gating had to be added to 2D convolutional layers as well. Considering that weights are shared across neurons in such layers typically, the dendritic/context weights for each neuron were assumed to be the same as well. This caused whole feature maps (i.e. channels) to be up or down-regulated at a time conditioned on the given context vector.

\cite{beyond} tested their ADNs without the context gating, leaving just sparse networks. These nets had only activation sparsity however, not weight sparsity. I tested sparse networks as well, because \cite{beyond} found that activation sparsity alone was better than dendritic segments alone.

\begin{figure*}
  \includegraphics[width=\textwidth]{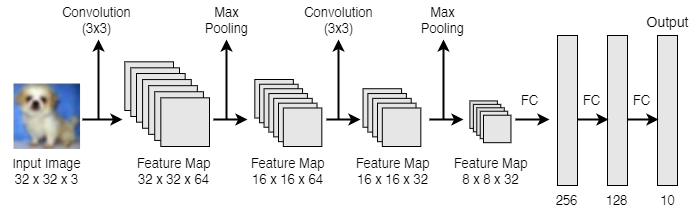}
  \caption{This is the base CNN used for experiments on SplitCIFAR100. Convolutional and fully connected layers are followed by the ReLU activation function for the base CNN shown, and kWTA for the ADNs and sparse networks. The ADNs had either task specific 1-hot encoded vectors, vectors full of 0s, or the task averaged input image flattened as a context signal at each convolutional and sparsely connected feedfoward layer. The hyperparameter grid search axes and resulting values are in \textit{\cref{appendix}}.}
  \label{lenet5}
\end{figure*}

Both the convolutional ADNs and sparse networks replaced ReLUs in \cref{lenet5} with kWTA activation functions, and fully connected layers with sparsely connected ones. The k-winners were taken across all output channels. The convolutional ADNs also had channel-wise context gating for the neurons in the convolutional layers.

Following from the results of \cite{dense, beyond, numenta_fpga}, I constructed grid searches as shown in \cref{appendix}. Their papers reported no impact of convolutional weight sparsity, so I did not test it further. The grid searches for the sparse networks and ADNs tested 12x as many hyperparameter sets relative to the grid search over the base CNN (see \cref{lenet5}).

%%%
\section{Results}
\label{Results}
%%%

\subsection{Iterative Magnitude Pruning Results}

\setlength{\belowcaptionskip}{-15pt}
\begin{figure}
  \centering
  \includegraphics[width=0.9\columnwidth]{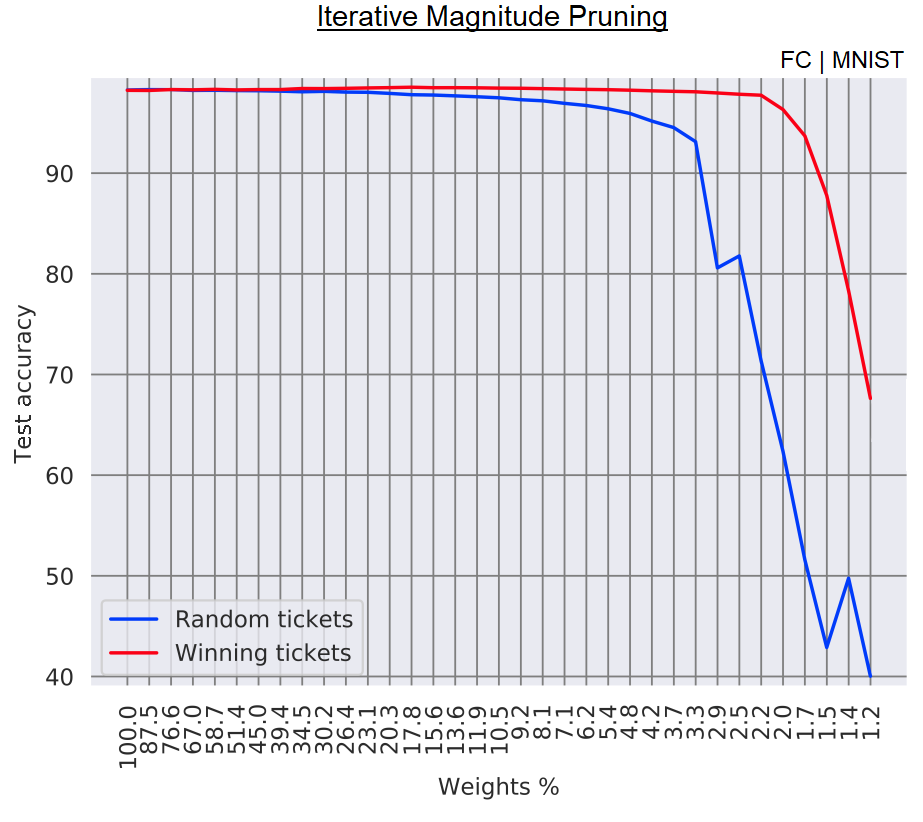}\\
  \vspace{-0.2cm}
  \includegraphics[width=0.92\columnwidth]{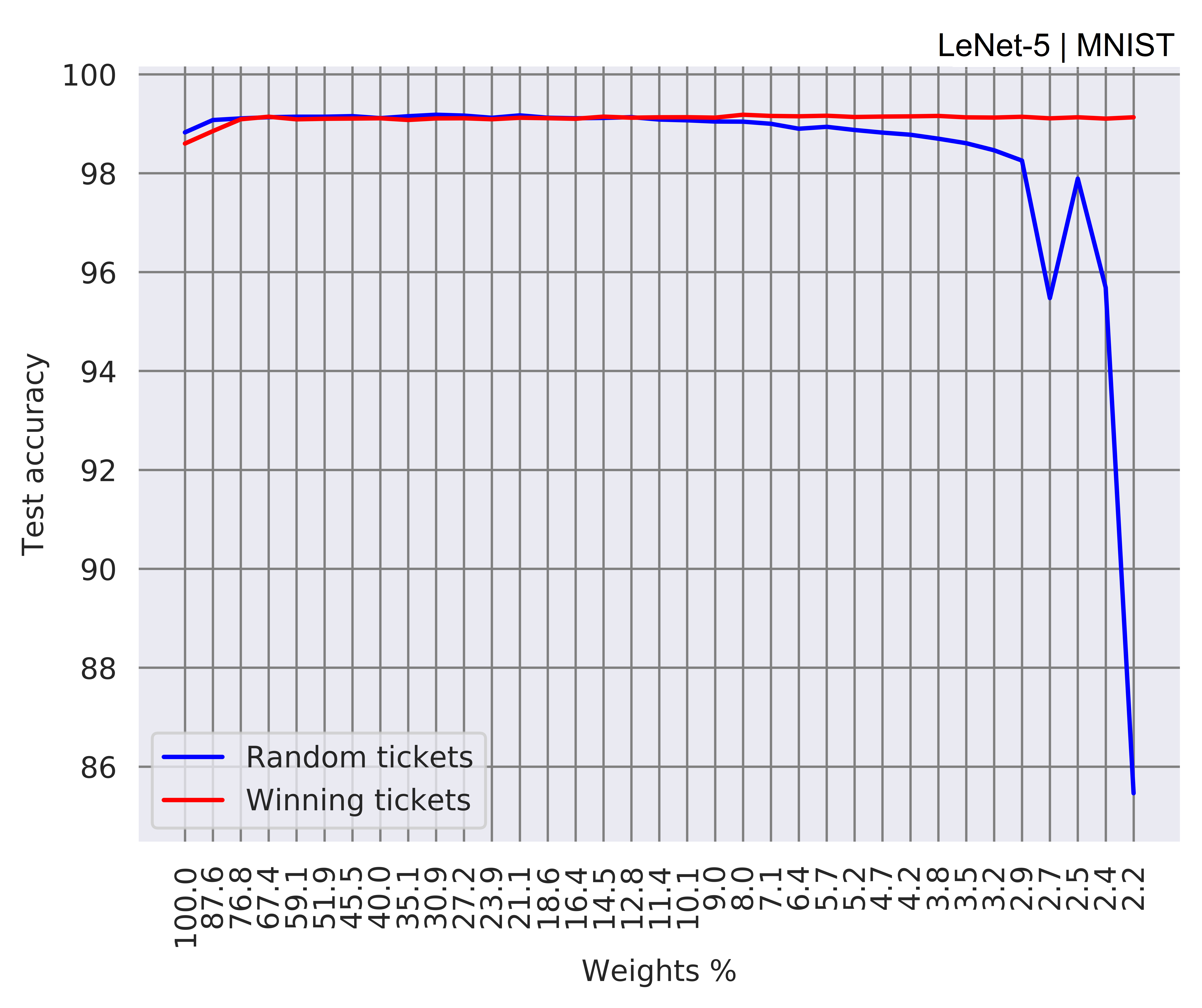}\\
  \vspace{-0.15cm}
  \includegraphics[width=0.9\columnwidth]{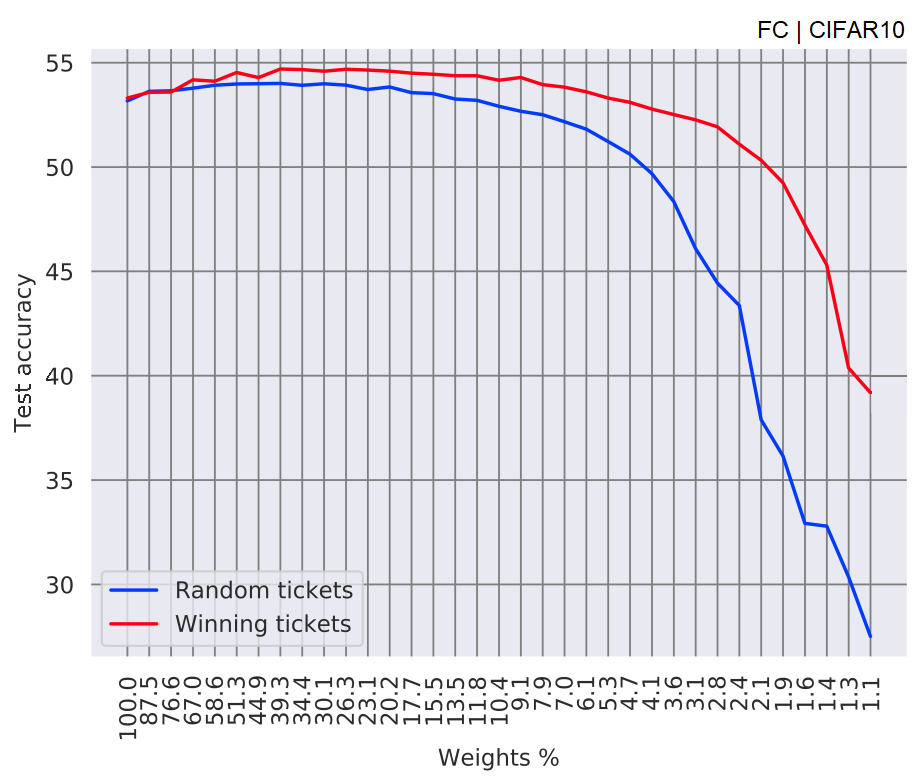}
  \caption{Averaged training curves from 8 randomly pruned networks and 8 winning lottery tickets produced with IMP (see \cref{pseudocode_imp}). 12.5\% of the remaining weights are pruned after each training iteration.}
  \label{fig:lth_train}
\end{figure}
\setlength{\belowcaptionskip}{0pt}

\cref{fig:lth_train} shows the test accuracy evolving with successive train-prune loops, and contrasts them to randomly pruned networks of the same sparsity. WLTs are attained until around 3\% sparsity, and random tickets even test as well as the fully dense network until around 30\%.\footnote{The middle graph's x-axis is slightly different. Do not know why, could be a trivially small rounding error at each pruning step as the first round of pruning removed 12.4\% instead of 12.5\%.}

\begin{figure}
  \centering
  \includegraphics[width=\columnwidth]{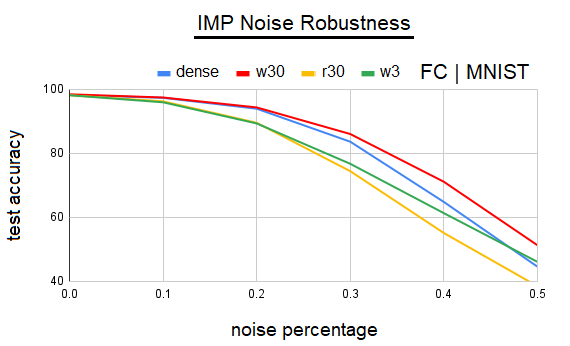}\\
  \vspace{0.35cm}
  \includegraphics[width=\columnwidth]{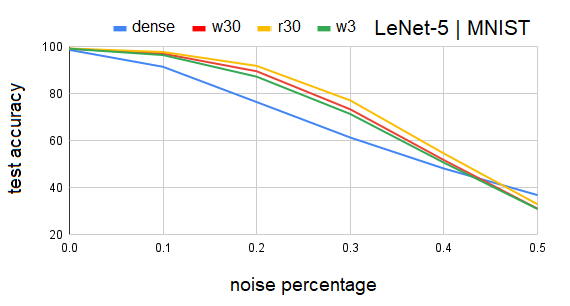}\\
  \vspace{0.2cm}
  \includegraphics[width=\columnwidth]{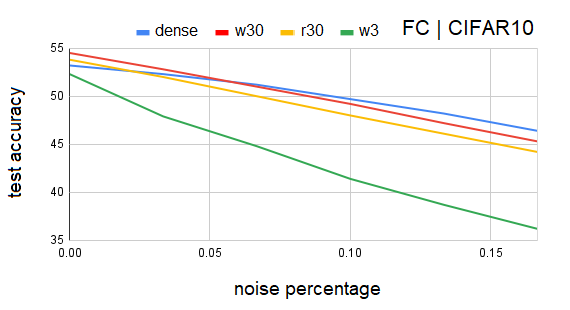}
  \caption{The effects of image noise on randomly pruned networks, winning lottery tickets, and fully dense networks. \textit{r30} are randomly pruned networks with 30\% of their weights remaining wrt the fully dense networks. \textit{w3} and \textit{w30} are winning tickets with 3\% and 30\% weights remaining respectively. Results are an average over 8 trained networks of each type.}
  \label{fig:lth_noise}
\end{figure}

Given this, WLTs of 3\% and 30\% sparsity, random tickets of 30\% sparsity, and dense with 100\% sparsity were tested for noise robustness by injecting uniform noise into all the test set images. The results in \cref{fig:lth_noise} show that there is no consensus as to which network does best. w30 does best on MNIST when using fully connected networks, and 2nd best on the other setups. Besides that, there is not much of a trend.

\subsection{Active Dendrite Network Results}

\begin{figure*}
  \includegraphics[width=\textwidth]{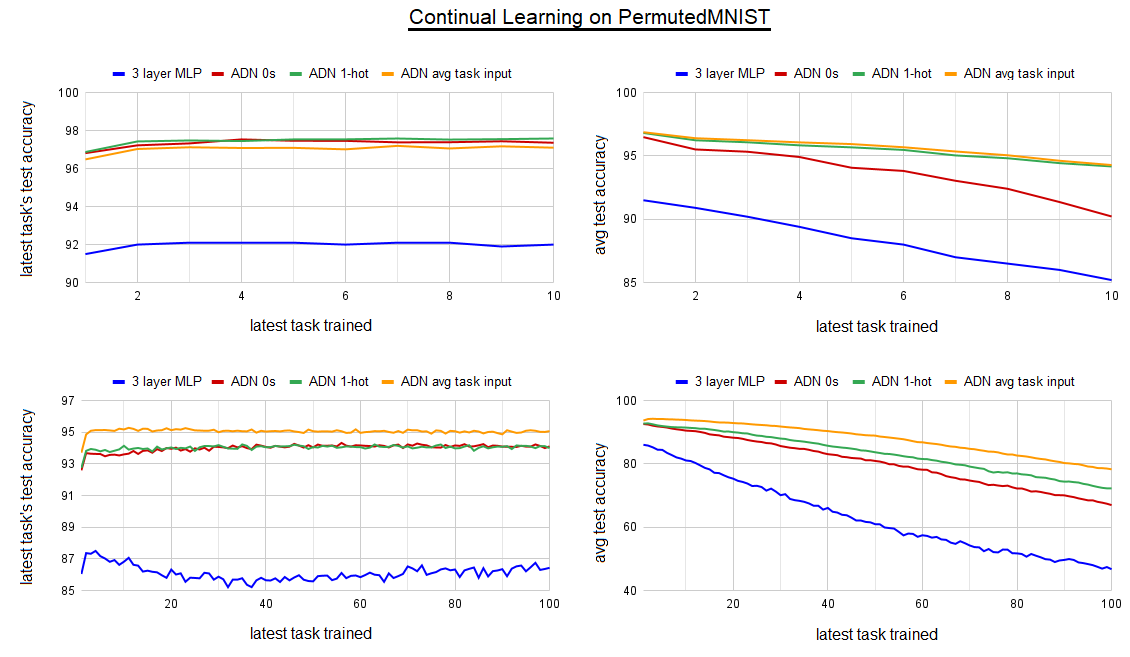}
  \caption{Replicated subset of results from \cite{beyond}. Additionally evaluated the importance of the context type. \textit{ADN 0s} has the same amount of dendritic weights as \textit{ADN 1-hot}, but always gives a context signal of 0s, essentially proving no task specific information. \textit{ADN avg task input} receives the task averaged input image flattened as context, which results in a much larger network (see \cref{context_gating} and \cref{adn_pmnist} for an explanation).}
  \label{replicated}
\end{figure*}

My experiments were able to replicate a subset of Numenta's \cite{beyond}. Specifically, their 10 segment ADN network got 94.6\% and 78.5\% on the 10 and 100 sequential tasks of PermutedMNIST respectively. This work's ADN networks got 0.3\% and 0.2\% less respectively. \footnote{They always tested on the same training dataset seed, including their hyperparameter sweep, and thus the same set and order of permutations. This most likely accounts for the minor performance differences.} 
The performance difference between the 1-hot and average image ADNs was small for the 10 task setting. The ADNs with 0s context had the majority of improvements wrt the baseline 3 layer MLP. Perhaps doing a separate hyperparameter search on the 100 task setting would reduced the performance gap between the 3 ADNs even more.

\begin{figure*}
  \centering
  \includegraphics[width=0.9\textwidth]{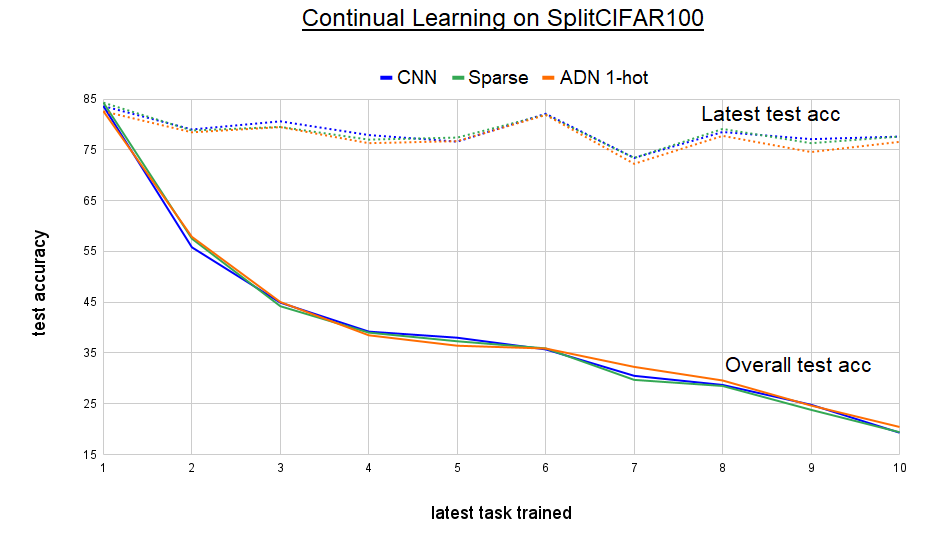}
  \caption{Latest and average task test accuracy while training on SplitCIFAR100, a dataset of random 10-way classification tasks made from CIFAR100. The hyperparameter grid search axes and resulting values are in \cref{appendix}}
  \label{fig:splitCifar}
\end{figure*}

The final results for SplitCIFAR100 are presented in \cref{fig:splitCifar}. The trends are nearly overlapping each other, indicating the base CNN, sparse version, and convolutional ADN all perform similarly well.

\section{Discussion}
\label{Discussion}

It was suggested by \cite{dense} that perhaps weight pruning techniques increase noise robustness. This work suggests that weight sparsity alone in FC and small CNNs does not improve noise tolerance in image classification scenarios. Perhaps the benefits \cite{dense} found were from activation sparsity alone, or from combination with weight sparsity. A seemingly easy way to test this would be to add a kWTA layer to the networks tested in this paper to see if noise robustness increases.

The epochs at each train-prune loop were not tracked during IMP, but the test accuracy rose very quickly at the beginning of all subsequent training phases (except for some of the more extremely pruned ones). I also played around with pruning earlier than convergence (e.g. after only 1 epoch of training), and it did seem to produce WLTs as well. This might be an easy way to reduce the amount of compute this method takes to work. In another paper, I explored the shape of the loss landscape using dimensionality reduction techniques. The shape of the loss wrt the weight/configuration space is fairly stable after ~100 examples \cite{bain_wlt}. Perhaps this phenomena can be used to motivate when pruning should occur.

The results on PermutedMNIST indicate that the great majority of the benefits can be had from multiplicative gating alone. I would be curious to know if learning a single multiplicative factor achieves the majority of continual learning benefits observed here and in \cite{numentas_latest_RL, beyond}.

The primary motivation for extending ADNs to CNNs came from the fact that the number of trainable parameters explodes. \cite{beyond} trained a rather shallow ADN with over 300 million weights to do a variant of MNIST. While the number of training parameters is large, they pointed out that some post-processing can be done to reduce the parameters to just 1 per active dendrite neuron. As mentioned in the \cref{dl_back}, learning fewer parameters was one of the original motivating factors for sharing convolutional weights. Nowadays we have big data and big compute. It might be time to revisit the biological plausiblity of weight sharing in striate cortex. DNNs in the past decade have benefited from removing weight sharing from convolutional layers \cite{no_share1, no_share2}.\footnote{Yes, I know we won't be guaranteed translational invariance. Perhaps the weights of some feature maps could remain shared.}

Alexnet made sparks in the community with only 2 GPUs totalling 6GB of VRAM \cite{alexnet}. Techniques today spend tens of thousands of dollars on renting compute, or sometimes millions on renting or building super computers (eg GPT-3, alphastar). The amount of compute thrown into searching the weight space is orders of magnitude larger nowadays \cite{hardware1, hardware2}. 

\cite{rate_code_og} in the 1920s found that hanging heavier weights from a muscle led to faster firing rates from neurons inside the muscles. This is where rate coding originated from, the idea that information in nervous systems are communicated in the form of average firing rates. This is surely one form of communication, and usually assumed to be what the ReLU activation function is modeling, yet several other temporal codes are known to be used in BNNs \cite{codes}. As evidence that DNNs use rate coding are the numerous successful ANN-to-SNN \footnote{Spiking Neural Networks, often thought to model more tricks from the brain} conversions methods rely on replacing activation values in DNNs (e.g. ReLU, sigmoid, and tanh) with rate-based coding \cite{ratebased1, ratebased2, ratebased3, ratebased4, ratebased5, ratebased6}. Many of these approaches suffer little accuracy loss from the conversion process, and can be implemented much more efficiently on neuromorphic hardware \cite{ratebased4, ratebased5, ratebased6, neuromorphic1}.

Numenta \cite{numenta_fpga} used FPGAs to benefit from both weight and activation sparsity. Combined with smaller network memory usage, the net effect was a 112x throughput increase on google's speech command dataset. Modern GPUs benefit from dense computations, and do not have very efficient random access memory relative to FPGAs \cite{numenta_fpga}. % "Most systems perform best on dense computations, where the predictability of memory access patterns allows data to be prefetched in a timely manner, and the dense data packing enables a processor's vector units to be leveraged to full effect" "Random access to memory is far more granular and efficient on an FPGA, enabling FPGA implementations to efficiently andle the unstructured access patterns in sparse networks."

Numenta \cite{dense} showed that their ADN network learned to activate sparse sub-networks, which facilitated retaining previously learned task skills. While this attempts to make DNN updates more like the brain's, it remains to be demonstrated how the multiplicative contextual gating is implemented in terms of BNNs. The literature on NMDA spikes point to different computations that interact in a non-WTA way, and instead they should combine their ability to produce more APs. Note that smaller dendritic spikes frequently fail to propagate to the soma, and instead rely on combining their effects with other spikes or bAPs to significantly affect the voltage at the soma \cite{helper1, helper2, helper3}. It is worth taking into account that DNN's uses rate coding and NMDA spike duration is linear wrt input strength, while imagining what effect dendritic spikes have on neuronal computations. In the future, someone should test 1D convolutions with shared weights as a way to model NMDA spikes since there are overlapping segments of dendrite that can trigger such spikes. Direction and a decaying sum could also be used to represent the spikes starting at a distal point of the dendrites and working their way towards the cell body, allowing the more distant signals to influence the more proximal ones.

Chavlis and Poirazi's recent work \cite{chavlis_21} also focuses on the potential benefits of modeling active dendritic processes in DNNs. One area they touch on that this work does not is the plasticity rules that govern the dynamic rewiring of synapses (i.e. connections) on dendritic arbors. They believe such rules could improve transfer and continual learning performance. This is a promising future direction given that most neural net architectures are static in their connections, where it is estimated that tens of thousands of connections are being removed or added every second in the adult human brain \cite{adult_syn}, and at least an order of magnitude more during the first 7 years of childhood \cite{child_syn}.

Nearly all mammals studied have abundant amounts of pyramidal neurons. They constitute perhaps the most important cell type in our brains, and are found primarily in areas associated with advanced cognition \cite{spruston08}. The literature shows that they have not yet been accurately modeled at the level of spike timing whether in SNNs or ANNs \cite{naud1, naud2, naud3}. 

Recently \cite{shit_oracle} used a detailed compartmental model to create a synthetic dataset of pyramidal neuron's spiking behavior, and trained a deep CNN to replicate them. They concluded that a network between 5 or 8 layers deep was sufficient to reproduce the spiking dynamics of a single real pyramidal neuron. While this would support the claim that our point neuron models are outdated, the fidelity of the oracle they used from \cite{oracle} is questionable. \cite{oracle} had data from 3 real L5b pyramidal neurons. They tried modeling spiking features of both the apical trunk and soma. They successfully created 4 models of 1 cell with a lot of manual intervention. They could not get their method to work for the other 2 cells of which they had data on. Using this modeling method to generate a dataset of 1,000s of synthetic spike trains seems unjustified.

%%%
%%% EVOLUTIONARY PRECAUTIONS: Stay on the path.
%%%

\subsection{Let's Stay On The Path.}

DNNs still struggles with the following task domains:

\begin{itemize}
    \itemsep-0.5em 
	\item Few shot learning
    \item Continual learning
    \item Goal alignment
    \item Common sense
    \item Natural language inference
    \item Sensor fusion
    \item Adversarial robustness
    \item Task \& motion planning
    \item Network architecture search (NAS)
\end{itemize}

Our brains have solutions to all of them, and runs on 20 watts \cite{andy_z}. \footnote{Even under seemingly demanding tasks it shows fairly constant usage. Lee Sedol was likely using 25 watts playing AlphaGo, which used ~6,800x more. A good portion of AlphaGo's algorithm is attributed to reinforcement learning's TD models from Sutton and Barto, who were modeling behavioral data from mammals (eg Pavlov's dogs).} Humans are often the implicit, yet appropriate, benchmark for artificial general intelligence (AGI). We are the only somewhat generally intelligent things that we can be certain exists.  Evolution by natural selection has spent astronomical compute and time discovering some really neat tricks. Some are inside ANNs today, but they could surely use more. We should continue following the path that evolution has already laid down. It is easier to reverse engineer something than it is to reinvent it, even something as complicated as the brain.

%\footnote{Ideas like these come from dual-inheritance models of evolution, sociobiology, ethology, evolutionary psychology, and memetics. Darwin knew of the pushback his insight would cause, but it is worth following it all the way to the conclusion that humans are still evolving. Sure, we can often have the most rapid, and largest component of the way things evolve over time, but that does not mean that we don't select on each other in such away that all favoritism washes away.}

% \clearpage
{\small
\bibliographystyle{ieee}
\bibliography{egbib}
}

\clearpage
%\newpage
\appendix
%\onecolumn

\section{Experimental Details}
\label{appendix}

\subsection{LTH Experiments}
\noindent
The LeNet-5 architecture used 3x3 kernels, max pooling, and ReLU activation functions after all 3 fully connected layers. The first convolutional layer took in a 1 channel image and outputs 64 feature maps. The latter convolutional layer takes in and outputs 64 channels.
\\
\noindent
The fully connected nets are the same for MNIST and CIFAR10, except for the variable length flattened input image dimensions: 784 and 3,072 respectively. The first fully connected layers output 300 neurons, and the final layer output 10, one for each output class.

\subsection{LTH Hyperparameters}
\noindent
All LTH experiments used the adam optimizer with a weight decay constant of 1e-4 and learning rate (LR) of 1.2e-3. Prune percent per training loop was 12.5\%. Batch size was 60.

\subsection{ADN Experiments}
\noindent
The adam optimizer was used for all experiments. All parameters of models used to replicate Numenta's \cite{beyond} previous findings were taken from their hyperparameter sweeps (i.e. the same values they used).

% tablesgenerator.com
\begin{table}[H]
\begin{tabular}{|l|c|c|c|c|c|c|}
\hline
\textbf{Network} & \multicolumn{1}{l|}{\textbf{LR}} & \multicolumn{1}{l|}{\textbf{batch size}} & \multicolumn{1}{l|}{\textbf{weight decay}} & \multicolumn{1}{l|}{\textbf{conv activation sparsity}} & \multicolumn{1}{l|}{\textbf{ff activation sparsity}} & \multicolumn{1}{l|}{\textbf{ff weight sparsity}} \\ \hline
CNN              & 1e-3                                & 64                                      & 1e-5                                       & 100\%                                                  & 100\%                                                & 100\%                                            \\ \hline
Sparse           & 1e-3                                & 32                                     & 0                                          & 20\%                                                    & 10\%                                                    & 50\%                                               \\ \hline
ADN              & 1e-3                               & 128                                      & 0                                          & 20\%                                                     & 30\%                                                  & 50\%                                         \\ \hline
\end{tabular}
\label{tab:split_pms}
\end{table}

\subsection{ADN hyperparameter search}

\noindent
\underline{Base CNN}: LRs [1e-5, 1e-4, 1e-3],\newline\textcolor{white}{Base CNN:} batch sizes [32, 64, 128],\newline\textcolor{white}{Base CNN:} adam weight decays [0, 1e-5, 1e-4]
\\
\noindent
\newline\textcolor{white}{Ba } \underline{Sparse}: LRs [1e-5, 1e-4, 1e-3],\newline\textcolor{white}{Base CNN:}  batch sizes [32, 64, 128],\newline\textcolor{white}{Base CNN:}  conv activation sparsities [0.1, 0.2, 0.3],\newline\textcolor{white}{Base CNN:}  ff activation sparsities [0.1, 0.3],\newline\textcolor{white}{Base CNN:}  ff weight sparsities [0.3, 0.5, 0.7]
\\
\noindent
\newline\textcolor{white}{Base }\underline{ADN}: LRs [1e-5, 1e-4, 1e-3], 
\newline\textcolor{white}{Base CNN:} batch sizes [32, 64, 128], \newline\textcolor{white}{Base CNN:} conv activation sparsities [0.1, 0.2, 0.3], \newline\textcolor{white}{Base CNN:} ff activation sparsities [0.1, 0.3], \newline\textcolor{white}{Base CNN:} ff weight sparsities [0.3, 0.5, 0.7]

\end{document}